\documentclass{article}

\usepackage{amsmath}
\usepackage{microtype}
\usepackage{graphicx}
\usepackage{subcaption}
\usepackage{todonotes}
\usepackage{caption}
\usepackage{booktabs} 

\usepackage{hyperref}

\usepackage[accepted]{icml2020}

\icmltitlerunning{Controlling Computation versus Quality for Neural Sequence Models}

\begin{document}

\twocolumn[
\icmltitle{Controlling Computation versus Quality for Neural Sequence Models
}



\icmlsetsymbol{equal}{*}

\begin{icmlauthorlist}
\icmlauthor{Ankur Bapna}{goo}
\icmlauthor{Naveen Arivazhagan}{goo}
\icmlauthor{Orhan Firat}{goo}
\end{icmlauthorlist}

\icmlaffiliation{goo}{Google Research, Mountain View}

\icmlcorrespondingauthor{Ankur Bapna}{ankurbpn@google.com}

\icmlkeywords{Machine Learning, Conditional Computation, Natural Language Processing}

\vskip 0.3in
]



\printAffiliations{}  

\begin{abstract}
Most neural networks utilize the same amount of compute for every example independent of the inherent complexity of the input. Further, methods that adapt the amount of computation to the example focus on finding a fixed inference-time computational graph per example, ignoring any external computational budgets or varying inference time limitations. In this work, we utilize conditional computation to make neural sequence models (Transformer) more efficient and computation-aware during inference.
We first modify the Transformer architecture, making each set of operations conditionally executable depending on the output of a learned control network. We then train this model in a multi-task setting, where each task corresponds to a particular computation budget. This allows us to train a single model that can be controlled to operate on different points of the computation-quality trade-off curve, depending on the available computation budget at inference time. We evaluate our approach on two tasks: (i) WMT English-French Translation and (ii) Unsupervised representation learning (BERT). Our experiments demonstrate that the proposed Conditional Computation Transformer (CCT) is competitive with vanilla Transformers when allowed to utilize its full computational budget, while improving significantly over computationally equivalent baselines when operating on smaller computational budgets.
\end{abstract}

\section{Introduction}
\label{sec:intro}
Over the last few years, scaling neural networks has tremendously improved the quality of models on several machine learning tasks. State-of-the-art Natural Language Processing models have billions of parameters, especially for tasks like Machine Translation \citep{shazeer2018mesh,huang2019gpipe}, Language Modeling \citep{radford2018improving} and Natural Language Understanding \citep{Devlin2018BERT,raffel2019exploring}. While training these models is feasible given the dramatic increase in the efficiency of training hardware \cite{jouppi2017datacenter} and research into efficient model-parallelism \cite{shazeer2018mesh,huang2019gpipe}, the amount of computation that can be expended at inference is often limited. However, these huge networks are usually inflexible and offer little control over the amount of computation used on any example, independent of the complexity of the input or the available computation budget for inference. 

Conditional Computation based approaches allow training networks where certain sub-networks can be conditionally executed, based on discrete decisions (optionally) trained with the model \citep{spall1992multivariate,Bengio2013EstimatingOP}. These methods also offer the potential for more control over the computation expended by the model during inference, conditioned on example difficulty or the available computation budget.

\begin{figure}[h]
  \begin{subfigure}[b]{0.22\textwidth}
    \includegraphics[width=\textwidth]{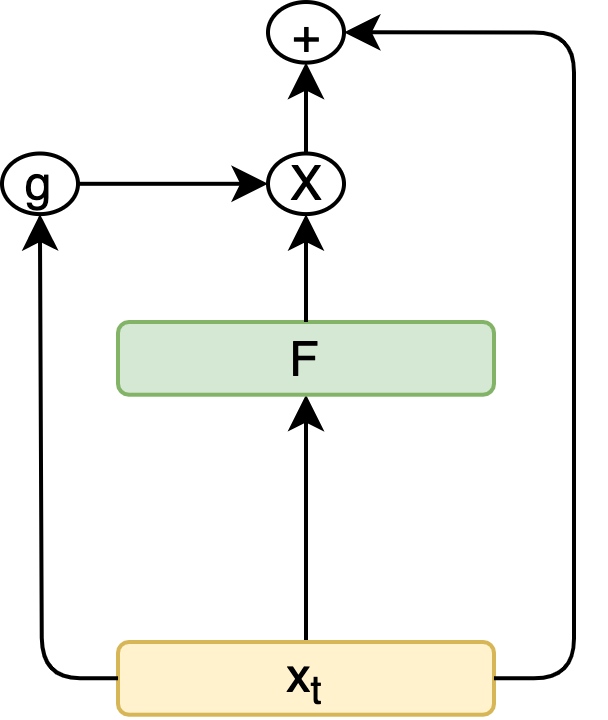}
    \caption{Training with noisy continuous gating.}
    \label{fig:f1}
  \end{subfigure}
  \hfill
  \begin{subfigure}[b]{0.18\textwidth}
    \includegraphics[width=\textwidth]{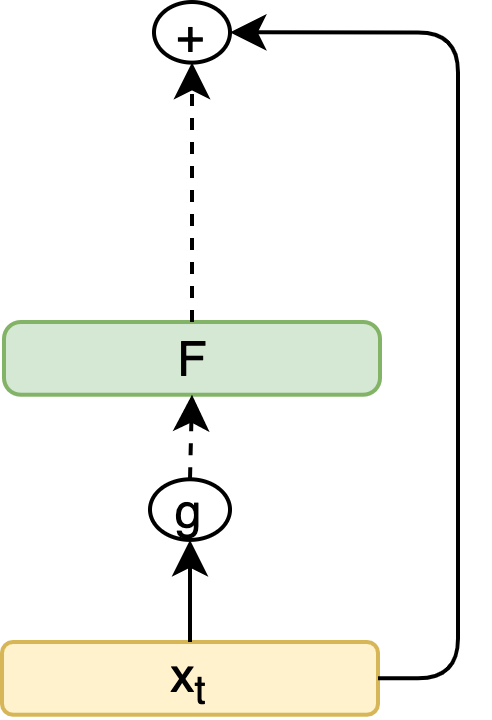}
    \caption{Inference with conditional execution.}
    \label{fig:f2}
  \end{subfigure}
  \caption{Our approach for adapting models for conditional computation: During training, sub-network outputs are gated by noised continuous outputs from control networks trained end-to-end with the model. During inference, sub-networks are conditionally executed depending on discrete outputs from control networks. Outputs are optionally short-circuited with residual connections.}
\end{figure}

Training a model with discrete intermediate outputs requires back-propagating through discrete random variables, which hinders model trainability. While several approaches have been suggested to alleviate this problem, including the use of gradient estimators \cite{Bengio2013EstimatingOP,jang2016categorical}
 and reinforcement learning \cite{bengio2015conditional}, training neural networks with conditionally executable sub-networks is still rare. As a consequence, most work involving conditional computation is restricted to very specific applications.
 
In this work we present a general framework to adapt models for conditional computation and control the amount of computation used at inference. We make three major contributions: (i) We provide a simple approach to adapt models for conditional computation by adding control networks trained end-to-end with the model. These control networks produce continuous outputs during training which allows for back-propagation. During inference these networks act like binary stochastic units that control the execution of their respective sub-networks. (ii) We propose a multi-task training approach to train a single model at different computation budgets. This allows controlling the amount of compute expended by the model on any example at inference. (iii) We adapt the Transformer architecture for conditional computation and demonstrate the efficacy of our approach on two large scale sequence modeling tasks: WMT'14 En-Fr Translation and Representation learning with BERT.




\section{Method}
\label{sec:method}

\subsection{Adapting models for conditional computation}
\label{subsec:theta}

We adapt neural sequence models for conditional computation by allowing the model to selectively execute certain sub-networks of the computation graph, conditioned on the outputs of small control networks learned jointly with the model.

Let the input to a layer $l$ be a sequence $X = \{x_1, ... x_t, ... x_T \}$ of length $T$. 
Let the output of this operation be $z_t$, given by $z_t = F_l(x_t)$. In the presence of residual connections this can be re-written as $z_t = F_l(x_t) + x_t$.


We now introduce a control network $g_l$ to control the execution of layer $l$. While it is possible to train neural networks with back-propagation in the presence of discrete outputs we preclude this problem by training in expectation. At training time, instead of sampling a discrete decision from $g_l(x_t)$, we compute the expected, $z^c_t = g_l(x_t)~F_l(x_t) + x_t$. We define the operation of $g_l$ by $g_l(x_t) = \sigma(G_l(x_t))$, where $G_l$ can be any function mapping $\mathcal{R}^d \rightarrow \mathcal{R}$ and $\sigma$ is the logistic sigmoid function. As a result $g_l(x_t) \in (0, 1)$ for any $x_t \in \mathcal{R}^d$ during training.

The gated version of layer $l$ can be written as:
\begin{equation}
    z^c_t = F^c_l(x_t) = g_l(x_t)~F_l(x_t) + x_t.
\end{equation}
At inference, this operation simplifies to
\begin{equation}
    z^c_t = \begin{cases}
            \begin{aligned}
            F_l(x_t) + x_t, \quad g_l(x_t)>= 0.5\\
            x_t, \quad  g_l(x_t) < 0.5\\
            \end{aligned}
        \end{cases}
\end{equation}

This, however, introduces a discrepancy between training and inference, with the former using soft decisions and the latter selectively executing layers based on discrete decisions. To bridge between these two modes of operation we encourage $g_l$ to become more discrete as training progresses. We follow the approach used in previous work for training binary stochastic neurons for monotonic attention mechanisms \citep{pmlr-v70-raffel17a,chiu2017monotonic,arivazhagan2019monotonic}.
We add zero-mean Gaussian noise to the output of $G_l$ during training, as shown below:
\begin{equation}
    \label{eq:noise}
    g_l(x_t) = \sigma(G_l(x_t) + \alpha ~\mathcal{N}(0, 1) ),
\end{equation}
where $\alpha$ increases linearly during the training process. This increasing schedule carries the pre-activation towards the saturation range of gating function $\sigma$ and in return, forcing the output of $g_l$ closer to the boundaries of $(0, 1)$. While $G_l$ could have any possible parameterization, for the purpose of this work we restrict it to single hidden layer feed-forward networks for simplicity.
\begin{equation}
    G_l(x_t) = RELU(x_t~W_1 + b)~W_2.
\end{equation}

\subsection{Modulating the Inference Budget}
\label{subsec:lambda}
In the absence of any other training signal, we would expect the training loss to pull the model towards using all (or most) of its computation in order to maximize performance. To control the amount of computation utilized by the model we impose a computational budget loss in addition to the training objective.

For any layer $l$, the expected cost of computation utilized by the model on any token, $x_t$, can be given by $g_l(x_t)~C_l$, where $C_l$ is the cost of applying layer $l$ to one token. For the purpose of this work we define the cost of a layer to represent its computational cost in terms of Flops.

Given a batch $B$ of sequences with $T$ time-steps, let $x_{t}^{(b)}$ be the $t$-th token of the $b$-th sequence. We define the computational budget $C_{budget}$ as a fraction, $p \in [0, 1]$, of the maximum computation available for the batch. Then the computational budget on the given batch of tokens is:
\begin{equation}
C_{budget} = p ~\Sigma_{b=1}^{|B|}\Sigma_{t=1}^T \Sigma_{l=1}^L C_l
\end{equation}
The expected computation used by the proposed conditional computation model is determined by the activations of the control networks on individual tokens of the batch: 
\begin{equation}
C_{util} = \Sigma_{b=1}^{|B|}\Sigma_{t=1}^T \Sigma_{l=1}^L g_l(x_t^{(b)}) C_l
\end{equation}

Then we define the computational budget loss on the given batch of tokens to be:
\begin{equation}
    \label{eq:cb}
    \mathcal{L}^c_B = \frac{1}{C_{budget}}|C_{budget} - C_{util}|
\end{equation}


We impose a constraint on the total computation used for a batch, instead of the compute used for a single sequence or token. This looser constraint allows the model to allocate more computation for `difficult' examples by using less compute for `easy' examples. Empirically we find that using the batch-level constraint performs better, especially at lower computation budgets.

Training a conditional computation model with the above loss allows operating that model at a single computation budget, $p$. Given a set of desired computational budgets, $P = \{p_1, p_2, ... p_N\}$, that we want the model to operate at, we utilize a simple multi-task training approach. We define a set of control symbols, $S = \{s_1, ... s_N\}$, which can be fed as additional inputs to the model. We associate each budget, $p_i$, with a control input $s_i \in S$. Given a batch of training sequences, $B$, we (uniform) randomly assign each sequence to a budget in the set $P$. Let the batch of sequences assigned to budget $p_i$ be $B_i$. The corresponding control symbol $s_i$ is then fed to the model as an additional input when training on sequences in $B_i$. By associating specific control inputs with different computational budgets, we train a single model to operate at specific levels of computation controlled by these external inputs. This is similar to approaches used for training multilingual Machine Translation models \citep{johnson2017google} and other multi-task models.

The total budget loss function in this multi-task training setup is then:
\begin{equation}
    \mathcal{L}^c = \Sigma_{i=1}^N\mathcal{L}^c_{B_i}
\end{equation}

In certain cases it might be desirable to control the amount of computation spent on different sub-networks of the model independently. For eg., in auto-regressive seq2seq models there is an inherent difference in the mode of operation of encoder and decoder sub-networks. To control the budgets for $M$ sub-networks independently, our multi-budget formulation can be extended to allow $P_{M} = \{(p_{11}, \dots, p_{1M}), \dots, (p_{N1}, \dots, p_{NM})\}$. Each symbol, $s_i$ then maps to a tuple of budgets, $(p_{i1}, \dots, p_{iM})$, specifying the desired budget for each sub-network.

The generalized budget loss function can be described as:
\begin{equation}
    \mathcal{L}^c = \Sigma_{i=1}^N\Sigma_{j=1}^M\mathcal{L}^c_{B_{ij}}
\end{equation}
where $\mathcal{L}^c_{B_{ij}}$ is the $i$-th budget loss for sub-network $j$.

Given a model adapted for conditional computation following the approaches described above, controlling the inference time computation just requires feeding the right control input, $s_i$, corresponding to the desired budget $p_i$.

\section{Conditional Computation Transformer}
\label{sec:cct}
We now apply our approach to the Transformer architecture \citep{vaswani2017attention}. We follow the new transformer layout where layer normalization (LN) is applied to the input instead of the output.\footnote{Please refer to the `nda' layout as implemented in the Tensor2Tensor library \cite{tensor2tensor}.}


\subsection{Conditional Attention Layer}
\begin{figure}[h]
  \centering
\includegraphics[width=0.4\textwidth]{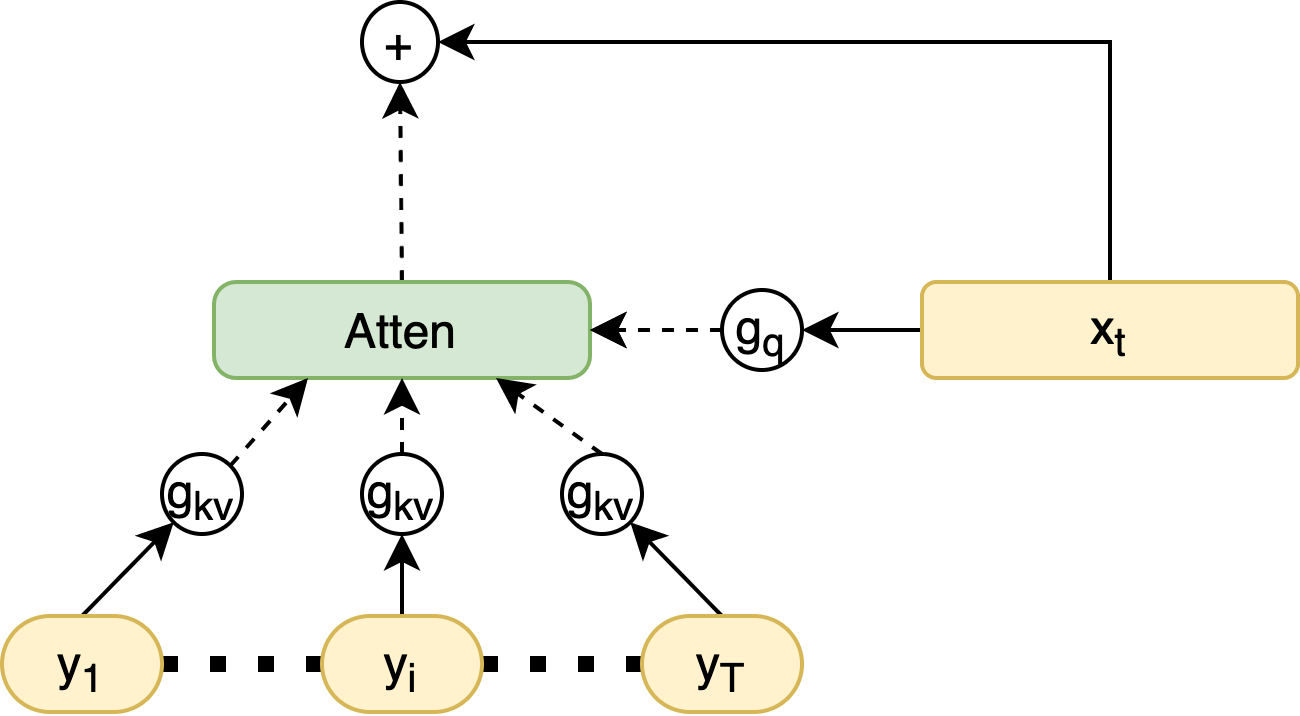}
  \caption{Conditional Computation Attention Layer.}
  \label{fig:wmt}
\end{figure}
Given a vector $x_t$ and a sequence of vectors, $Y = \{y_1, y_2, \dots ,y_T\}$,  the transformer attention layer can be described by the following sequence of operations. The set of vectors to be attended are first projected to keys ($K$) and corresponding values ($V$)
\begin{equation}
    \label{eq:kv}
    \begin{aligned}
    K = YW_k\\ V = YW_v
    \end{aligned}
\end{equation}
\noindent The projected input queries are then used to attend the keys to summarize the set to be attended. 
\begin{equation}
    \label{eq:q}
    \begin{aligned}
    q_t &= LN(x_t)W_q\\
    a_t &= MultiHeadAtten(K, V, q_t)\\
    z_t &= Dropout(a_t) W_o + x_t
    \end{aligned}
\end{equation}

\noindent We introduce two control networks to control the execution of the operations defined by Equations \ref{eq:kv} and \ref{eq:q} respectively. The first control network, $g_{KV}: \mathcal{R}^{Td} \rightarrow (0, 1)^T$, controls the execution of the key-value projections. The second network, $g_q:\mathcal{R}^{d} \rightarrow (0, 1)$, controls the execution of the query projection, multi-headed attention and the attention post-projection. We also introduce additional normalization layers to stabilize training in the presence of discrete operations. During training we implement these changes as:
\begin{equation}
    \label{eq:ckv}
    \begin{aligned}
    K^c = g_{KV}(Y)LN(YW_k)\\ V^c = g_{KV}(Y)LN(YW_v)
    \end{aligned}
\end{equation}
\noindent and
\begin{equation}
    \label{eq:cq}
    \begin{aligned}
    q_t &= LN(x_t)W_q\\
    a^c_t &= MultiHeadAtten(K^c, V^c, q_t)\\
    z^c_t &= g_q(x_t)Dropout(LN(a^c_t)) W_o + x_t
    \end{aligned}
\end{equation}

The above modifications are applied to all self-attention and cross-attention layers.

\subsection{Conditional Feed-forward Layer}
\begin{figure}[h]
  \centering
\includegraphics[width=0.4\textwidth]{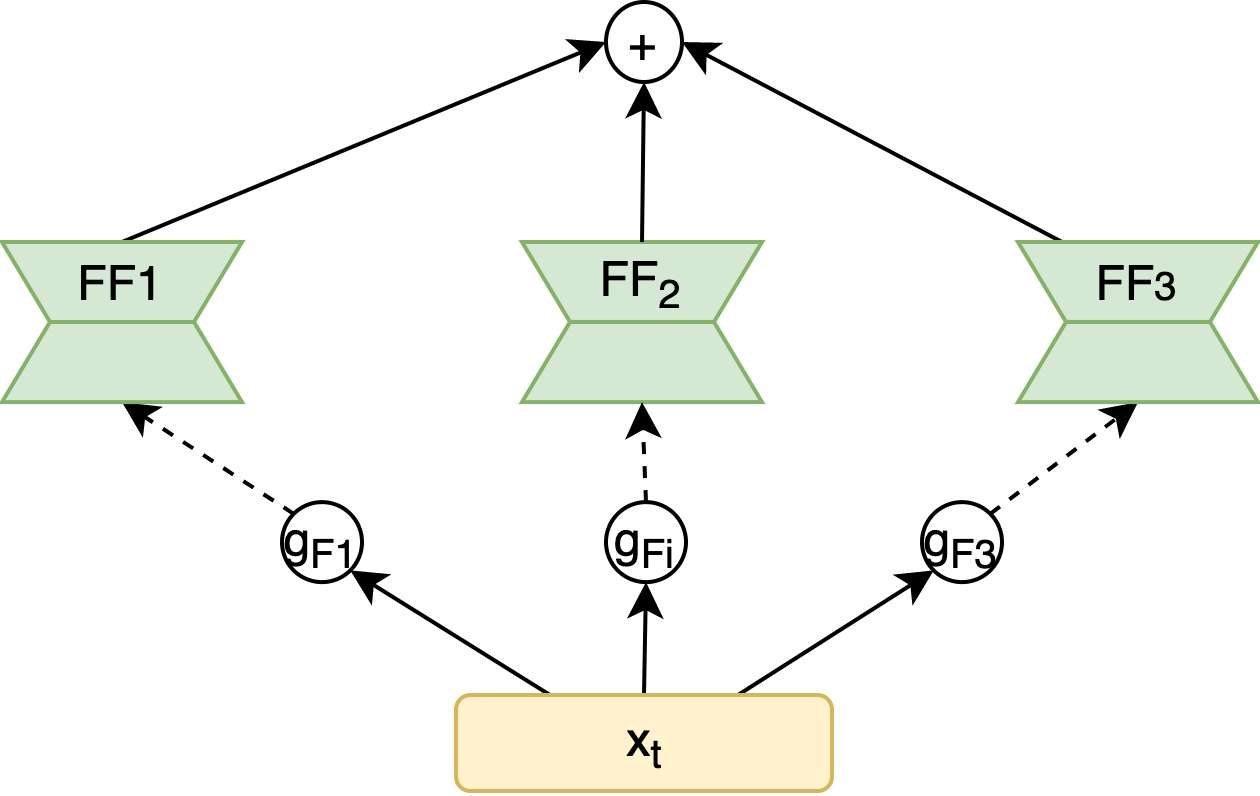}
  \caption{Conditional Computation Feedforward Layer.}
  \label{fig:wmt}
\end{figure}
Given a vector, $x_t$, the transformer feed-forward layer can be described by
\begin{equation}
    FF(W_1, W_2, x_t) = RELU(LN(x_t)W_1 + b)W_2    
\end{equation}
where $W_1 \in R^{d\times d_h}$ and $W_2 \in R^{d_h\times d}$. Then the output of the layer incorporating residual connections is given by $z_t = FF(W_1, W_2, x_t) + x_t$ .

Adding conditional execution for this layer, our output can be written as $z^c_t = g_{F}(x_t)LN(FF(W_1, W_2, x_t)) + x_t$.

While it's straightforward to add conditional execution for the entire feed-forward layer, we can optionally decompose the large feed-forward layer into independently controlled smaller layers to provide more granular control over feed-forward layer capacity\footnote{This decomposition makes our feed-forward layer similar to the Sparsely Gated Mixture-of-Experts layer \cite{shazeer2017outrageously}. However, in our approach the number of experts applied per input are a function of the input.}
\begin{equation}
    \label{eq:mff}
    \begin{split}
    z_t = \Sigma_{i=1}^Mg_{Fi}(x_t)LN(FF(W_{1i}, W_{2i}, x_t)) + x_t
    \end{split}
\end{equation}
where $W_{1i} \in R^{d\times \frac{d_h}{M}}$, $W_{2i} \in R^{\frac{d_h}{M}\times d}$ and $g_F$ maps the input to $(0, 1)^M$.

\subsection{Feeding Control Input}
\label{sec:feeding}
For training with multiple computation budgets, as described in Section \ref{subsec:lambda}, we need to feed an additional control input, $s_i$, with every input sequence. Given an input sequence, $X = \{x_1,\dots, x_T\}$, input to transformer layers is a sequence of embeddings corresponding to each input symbol summed with the corresponding position embedding. In addition to the position embeddings we learn an additional input embedding of control symbols, $S = \{s_1, ... s_N\}$. The embedding of the symbol $s_i$ is then added to the embedding of each token $x_t$ before feeding into the model.
\begin{figure}
  \centering
\includegraphics[width=1\linewidth]{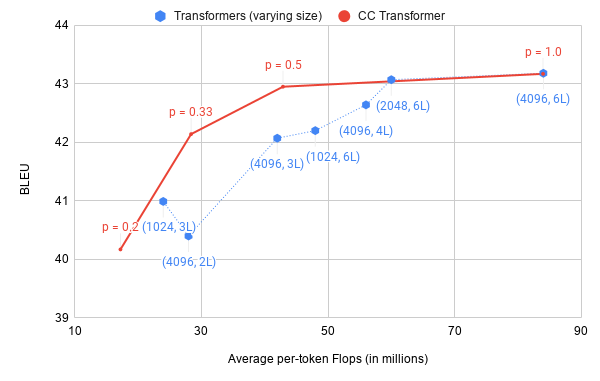}
  \caption{Comparing the performance of CCT (red) at different encoder-decoder computation budgets against Transformer baselines (blue). x-axis corresponds to the average encoder-decoder per-token Flops (in millions). The transformer network size is denoted next to each corresponding data-point using the format (hidden layer size, number of layers). Note: We do not compare the computation required for embedding lookup and softmax operations.}
  \label{fig:wmt}
\end{figure}

\section{Experiments on Machine Translation}
Most machine translation models fall within the sequence-to-sequence paradigm \cite{DBLP:journals/corr/SutskeverVL14, DBLP:journals/corr/BahdanauCB14}, with an encoder that learns representations of the source sequence and a decoder to generate the target sequence, trained on the cross-entropy loss $\mathcal{L}^{MT}$. Since there is a difference in the inference-time operation of the encoder and decoder (the encoder processes all source tokens simultaneously, while the decoder processes each token one at a time), we allow controlling their respective computation budgets separately. To elaborate, we permit using a set of computation budgets $P_{s2s} = \{(p_{1e}, p_{1d}), ... (p_{Ne}, p_{Nd})\}$. Here the first budget of every tuple, $p_{ie}$, corresponds to the desired encoder budget while the second budget, $p_{id}$, corresponds to the desired decoder budget. The control symbol, $s_i$, is fed as an embedding added to every source and target token. We train the model end-to-end on $\mathcal{L}^{MT} + \lambda~\mathcal{L}^c$

\begin{figure}
  \centering
\includegraphics[width=1\linewidth]{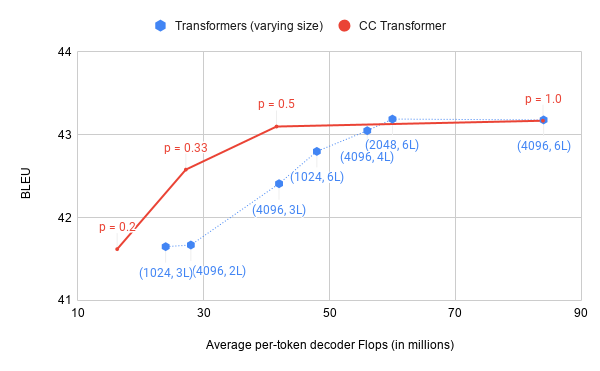}
  \caption{Comparing the performance of CCT at different decoder computation budgets against Transformer baselines, when allowed to use full encoder computation. x-axis corresponds to the decoder per-token Flops (in millions). Blue dots denote the quality of individual transformer baselines. The decoder size is denoted next to each corresponding data-point using the format (hidden layer size, number of layers). Note: We do not compare the computation required for embedding lookup and softmax operations.}
  \label{fig:wmt_dec}
\end{figure}

We now evaluate our approach on the WMT'14 English-French translation task\footnote{\url{ statmt.org/wmt14/translation-task.html}}. We use newstest13 for validation and newstest14 for test. BLEU scores are computed with tokenized true-cased output and references with Moses \textit{multi-bleu.perl}\footnote{\url{github.com/moses-smt}}.

We train a Transformer Big \cite{vaswani2017attention} model as our baseline. For smaller budget baselines, we reduce the capacity of our Transformer following two approaches: (i) Reducing the model depth by reducing the number of layers and (ii) Reducing the model width by reducing the hidden dimension of the feed-forward layers. We compare these baselines against a single CCT model operating at different computation budgets, with a maximum capacity equivalent to Transformer Big. Since decoder computation is typically the bottleneck for Transformer inference, we also train additional baselines where we only reduce the capacity of the decoder, while using a full Transformer Big encoder. These baselines are compared against the same CCT model from the previous comparison, but use full encoder computation while varying the decoder budget.

We use a Transformer learning rate schedule \citep{vaswani2017attention} of (3.0, 40K)\footnote{(3.0, 40K) schedule is the shorthand for a learning rate of 3.0, with 40K warm-up steps for the schedule, which is decayed with the inverse square root of the number of training steps after warm-up.} and all dropout probabilities are set to 0.1. For all our models, we use a shared vocabulary Sentence Piece Model \citep{kudo-richardson-2018-sentencepiece} for sub-word tokenization, with a vocabulary size of 32000 tokens. We train each model for 300k steps with batches of 250k tokens. The CCT is trained with the same set of hyper-parameters. In addition to the above hyper-parameters, we set $\lambda = 1.0$ and use a set of computation budgets, $P_{s2s}$, set to $\{1.0, 1.0, 1.0, 0.5, 0.33, 0.2\} \times \{1.0, 1.0, 1.0, 0.5, 0.33, 0.2\}$. This results in $36$ control tasks, one corresponding to each tuple from the above cross-product\footnote{We empirically find that allowing half of the control tasks to use their entire computational budgets strikes a good balance between properly training all parameters and learning to operate at reduced budgets.}. The noise factor, $\alpha$ in Equation~\ref{eq:noise}, is linearly increased during the training process, from $0.0$ at the first step to $5.0$ at 300k steps. For these experiments, we break the feed-forward layer into $4$ smaller layers i.e. $M=4$ in Equation \ref{eq:mff}. All our models are trained on 32 Cloud TPUv3 chips and evaluated at 300k steps.

Figure \ref{fig:wmt} compares a single CCT against individual Transformer models with different amounts of encoder and decoder capacity. Our results suggest that CCT is competitive with Transformer Big even when operating at half its computation budget. At smaller computation budgets CCT improves over smaller baseline Transformer models by up to 1-1.5 Bleu. Figure \ref{fig:wmt_dec} depicts the results of our second experiment, comparing CCT against baseline Transformers when using a large encoder (equivalent to Transformer Big) and controlling for decoder computation. We observe a similar trend, with CCT being competitive with Transformers at higher computation budgets, while improving over baselines by almost 1 Bleu at reduced budgets.

\section{Experiments with BERT}
\begin{figure*}[t]
  \centering
\begin{subfigure}[b]{\columnwidth}
\includegraphics[width=\columnwidth]{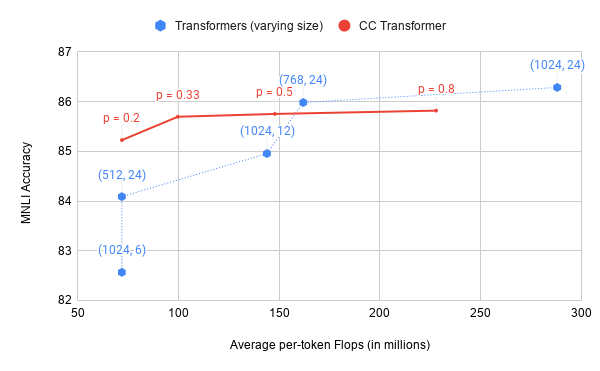}
  \caption{Comparison on MNLI validation set.}
  \label{fig:mnli}
\end{subfigure}
\begin{subfigure}[b]{\columnwidth}
  \centering
\includegraphics[width=\columnwidth]{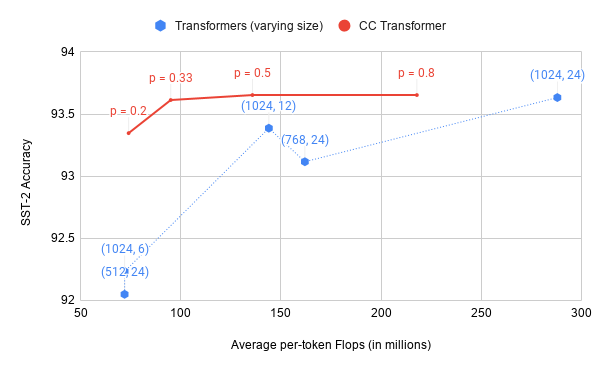}
  \caption{Comparison on SST-2 validation set.}
  \label{fig:sst}
\end{subfigure}

\begin{subfigure}[b]{\columnwidth}
  \centering
\includegraphics[width=\columnwidth]{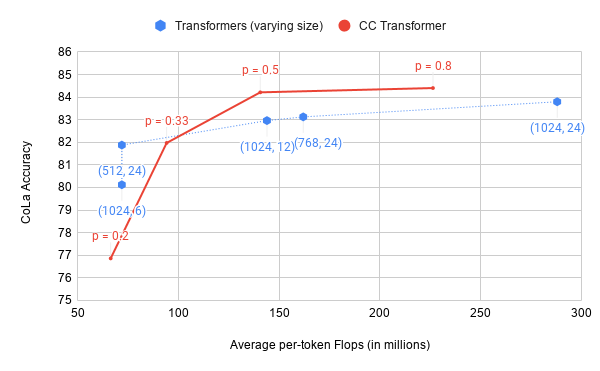}
  \caption{Comparison on CoLA validation set.}
  \label{fig:cola}
\end{subfigure}
\begin{subfigure}[b]{\columnwidth}
  \centering
\includegraphics[width=\columnwidth]{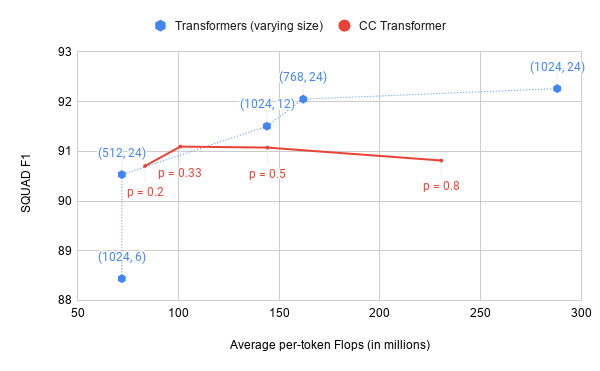}
  \caption{Comparison on Squad v1.1 validation set.}
  \label{fig:squad}
\end{subfigure}
  \caption{Comparing the performance of CCT at different encoder computation budgets against Bert baselines of different sizes. The transformer network size is denoted next to each corresponding data-point using the format (model size, number of layers).}
\end{figure*}
BERT \cite{Devlin2018BERT} uses a Masked Language Modeling objective, $\mathcal{L}^{mlm}$, in order to learn token-level representations of text using a Transformer architecture. Following the pre-training stage, the model is fine-tuned on individual tasks by training on (smaller) task-specific datasets and objectives. To control the amount of computation used by BERT for generating representations of text, we replace the Transformer model in the original BERT implementation with our CCT from Section \ref{sec:cct}. We train this model following the multi-computation budget recipe described in Section \ref{subsec:lambda}. The objective function used for training this model is $\mathcal{L}^{mlm} + \lambda\mathcal{L}^c$. When fine-tuning on a downstream task we use a different $\lambda$ scaled to the new objective.

We train a BERT-Large \cite{Devlin2018BERT} model as our baseline. For smaller budget baselines, we reduce the capacity of BERT following two approaches: (i) Reducing the model depth by reducing the number of layers and (ii) Reducing the width by reducing the model dimension and hidden dimension of the feed-forward layers, maintaining a ratio of 4 between the model dimension and feed-forward hidden dimension. We compare these baselines against a single CCT model operating at different computation budgets, with maximum capacity equivalent to BERT-Large.

We use the same pre-training process used in \citet{Devlin2018BERT}, except for one difference: we train on sequences of length 512 for 1M steps with a batch size of 1024, instead of training on shorter sequences for 900k steps and fine-tuning with longer sequences. The CCT is trained with the same set of hyper-parameters. In addition to the above hyper-parameters, we set $\lambda = 0.3$ and use a set of computation budgets $P_{BERT} = [0.8, 0.8, 0.8, 0.5, 0.33, 0.2]$. $\alpha$ is linearly increased during the training process, going from $0.0$ at the first step to $5.0$ at 300k steps and capping at that value. For these experiments, we break the feed-forward layer into $4$ smaller layers i.e. $M=4$ in Equation \ref{eq:mff}. All our models are trained on 64 Cloud TPUv3 chips.

When fine-tuning BERT baselines on downstream tasks, we search over the same grid used in \citet{Devlin2018BERT}. We re-use the \textbf{same} fine-tuning parameters as BERT-Large for fine-tuning CCT. The value of $\lambda$ used for fine-tuning CCT is different from pre-training, to scale to the downstream task loss. We report validation performance on 4 GLUE benchmark \cite{Wang2019glue} tasks over 3 runs: MNLI, SST-2, Squad and CoLA. A comparison of CCT with comparable baselines on MNLI, SST-2, CoLA and Squad tasks is depicted in Figures \ref{fig:mnli}, \ref{fig:sst}, \ref{fig:cola} and \ref{fig:squad} respectively. On MNLI and SST-2 we see a trend similar to translation, and the performance of CCT is close to the performance of BERT-Large at the highest computation setting, while improving significantly over baselines at smaller computation budgets. On CoLA we see the reverse trend: CCT improves by a significant margin at the highest computation setting while losing to baselines at smaller computation budgets.

The performance of CCT on Squad is worse than baselines at all computation budgets. It is worth noting that Squad is the only benchmark task that uses token level outputs from the pre-trained representations, while all other tasks act on a pooled representation of the entire sequence. The weak performance on Squad suggests that token-level representations extracted from CCT-BERT, without pooling, might not perform as well as those from a static architecture that uses the same set of operations for every token.

\section{Ablations}

\begin{figure}[t]
\includegraphics[width=\columnwidth]{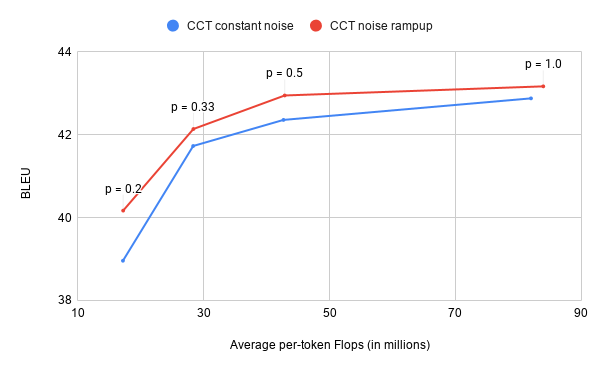}
  \caption{Comparing the performance of CCT when using linearly increasing Gaussian noise ($\alpha = 0.0$ at the first step, $\alpha=5.0$ at 300k steps) against using noisy discrete decisions from the beginning of training ($\alpha = 5.0$ for the entire process).}
  \label{fig:noise}
\end{figure}
\begin{figure}
\includegraphics[width=\columnwidth]{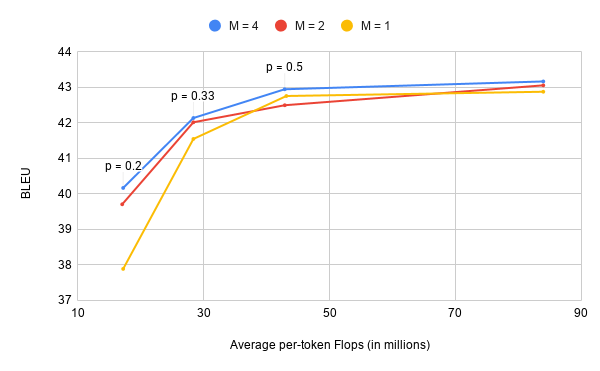}
  \caption{Comparing the performance of CCT when using different splits for the feed-forward layer ($M \in \{1,2,4\}$ in Equation \ref{eq:mff}).}
  \label{fig:mff}
\end{figure}

In order to shed more light on the factors effecting the quality and performance of the proposed CCT approach, we conduct further analysis probing various design choices.

\paragraph{Importance of the Noise Schedule} We next attempt to understand the role of gradually increasing noise variance when training with discrete decisions. We compare a CCT model trained with linearly increasing $\alpha$ (Equation \ref{eq:noise}) against one where it is set to its highest value (here $5.0$) from the beginning of training. From Figure \ref{fig:noise}, we notice that the two models are within 0.3 BLEU of each other at $p = 1.0$, with the difference increasing to 0.5 for $p = 0.5$ and $p = 0.33$. At $p = 0.2$, the performance of the discrete model deteriorates much faster with the difference growing to more than 1 BLEU. This suggests that the quality of control network training has a larger effect on model performance at smaller computation budgets.

\paragraph{Importance of Parallel Sub-Networks} Deciding how to divide the model's computation graph into sub-networks controlled by different control networks can have a significant impact on model quality. For example, for most of our experiments we split each feed-forward layer into 4 smaller, independently controlled feed-forward layers ($M=4$ in Equation \ref{eq:mff}). We compare the effect of splitting the feed-forward sub-network at different granularities (setting $M \in \{1, 2, 4\}$). Our results from Figure \ref{fig:mff} suggest that having more control on how network computation is utilized, by having control networks for smaller sub-networks, significantly impacts model quality especially at lower computation budgets.

\paragraph{Tricks of the trade}
We list some tricks and observations that were empirically found to be useful during the course of this work.

(i) Careful normalization was critical for stable training and good model quality. This includes additional layer normalization applied to every gated sub-network output and using separate layer normalization for the input of every independently gated feed-forward sub-network (i.e. separate layer-normalization for each of the $M$ feed-forward layers).

(ii) The range of budgets ($P$ from Section \ref{subsec:lambda}) used during training affected model quality. We observed significant quality deterioration when one of the values in $P$ was too low ($p < 0.05$). For BERT experiments setting $p = 1.0$ resulted in worse performance on the MLM loss at a budget of $p=1.0$ while performance at other budgets was not severely impacted. We suspect this is caused by the special role of the MASK token during pre-training.

(iii) Even with all the stabilization approaches, approximately $20\%$ of our runs deteriorated in performance at lower budgets on training further beyond convergence. We suspect this is a consequence of the multi-task training approach, resulting from different covergence rates at different budgets.

(iv) Varying control network capacity did not have a huge effect on model quality within the range of hidden dimensions evaluated by us ($\{64, 128, 256\}$).

(v) The proposed computational budget loss (Equation \ref{eq:cb}) is two-sided and also penalizes the model for under-utilization. While this is counter-intuitive, in practice we found that not penalizing the model for using less computation resulted in under-training certain sub-networks, resulting in sub-optimal downstream quality.

\section{Analysis}
The discrete nature of the control network outputs makes it possible to analyze which sub-networks are active for certain inputs under limited computation budgets. In order to understand the behavior of CCT, we analyze the activations of the control networks of our En-Fr machine translation model. For each sentence and reference translation, we first compute the control network activations for every token and then analyze the aggregate behavior over the entire validation set. We conduct this analysis on 3 independent training runs (with distinct initialization) and on both, the reference translations and the model's decoded translations with beam search. While the figures are based on the analysis conducted on the first run, we describe any observed qualitative differences across different runs in the main text.

\begin{figure*}[t]
  \centering
    \begin{subfigure}{0.9\columnwidth}
    \includegraphics[width=0.9\columnwidth]{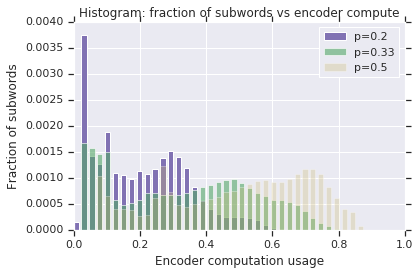}
    \end{subfigure}
    \begin{subfigure}{0.9\columnwidth}
    \includegraphics[width=0.9\columnwidth]{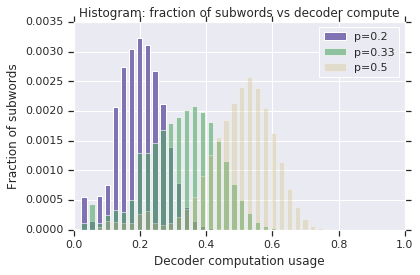}
    \end{subfigure}
  \caption{Left and right plots highlight the fraction of subwords using a certain fraction of the available encoder / decoder computation (normalized to between 0 and 1). The x-axis correspond to the fraction of encoder (left plot) or decoder (right plot) computation used by a sub-word, while the y-axis plots the fraction of sub-words falling within each bucket.}
  \label{fig:comp_subword}
\end{figure*}

\begin{figure*}[t]
  \centering
    \begin{subfigure}{0.9\columnwidth}
    \includegraphics[width=0.9\columnwidth]{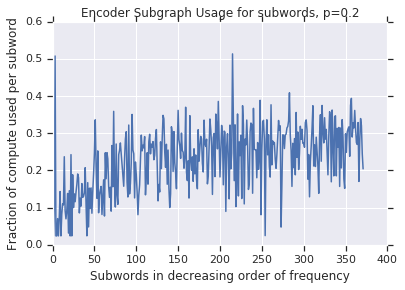}
    \end{subfigure}
    \begin{subfigure}{0.9\columnwidth}
    \includegraphics[width=0.9\columnwidth]{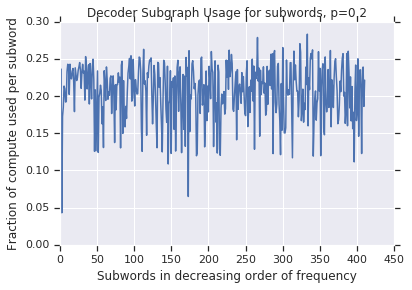}
    \end{subfigure}
  \caption{Left and right plots compare the fraction of available encoder (left plot) or decoder (right plot) computation used by a token, against the sub-word occurrence frequency in the validation set. The x-axis depicts subwords arranged in decreasing order of occurrence frequency, while the y-axis depicts the fraction of computation used.}
  \label{fig:comp_subword_freq}
\end{figure*}


\paragraph{Less computation for `easy' sub-words, more for others:} One approach to reducing the amount of computation utilized by a model is to utilize structured pruning based approaches, which prune out layers (partially \citep{voita2019analyzing,michel2019sixteen}, or entirely \citep{fan2019reducing}) from a larger model. Our approach equips the model with an input-dependent `self-pruning' mechanism instead. To illustrate how this approach allows the CCT to utilize computation more efficiently, we first evaluate the computation usage for every subword (token) in our validation set by evaluating their control network activations. We plot the spread of computation used across different subwords in the encoder and decoder of our En-Fr translation model in Figure~\ref{fig:comp_subword}. We find that the distribution of this spread is very different in the encoder and the decoder. The spread of computation in the encoder is bi-modal for all computational budgets, with one mode close to 0 and the other mode at a usage level much higher than the permitted budget.  We hypothesize that the availability of bi-directional context in the encoder allows encoding `easier' tokens with close to no compute (or encoding them jointly with neighboring tokens) while allocating more computation for more `difficult' tokens. On the other hand the computation spread in the decoder looks almost Gaussian centered on the budget $p_i$, suggesting that the auto-regressive, uni-directional nature of the decoder might require a certain minimum amount of computation to predict every output.
\begin{figure*}[t]
  \centering
    \begin{subfigure}{0.9\columnwidth}
    \includegraphics[width=0.9\columnwidth]{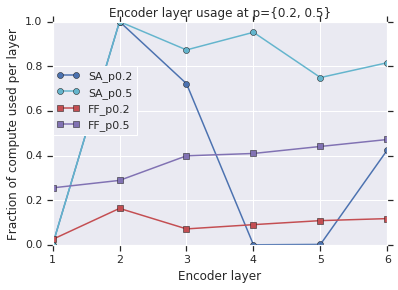}
    \end{subfigure}
    \begin{subfigure}{0.9\columnwidth}
    \includegraphics[width=0.9\columnwidth]{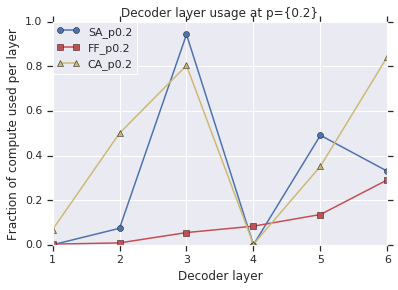}
    \end{subfigure}
  \caption{Comparing the fraction of $g_L=1.0$ (equivalent to the fraction of times a particular sub-network was active) for all layers in the encoder and decoder at different budgets. SA, FF and CA correspond to the self-attention, feed-forward (averaged over $M$) and cross-attention sub-networks respectively, while $p0.2$ and $p0.5$ indicate the computation budget.}
  \label{fig:capalloc}
\end{figure*}

\paragraph{Frequent sub-words require less compute to encode:} Given the huge spread in the amount of computation used per token, we next analyze if the computation required to encode or predict tokens has any relationship with their occurrence frequency. We evaluate the computation usage for every subword in the validation set, and average it over all occurrences of the subword. In Figure~\ref{fig:comp_subword_freq} we plot the fraction of encoder or decoder computation used per-subword against their frequency of occurrence in the validation set. We find that the amount of encoder computation used increases as the occurrence frequency of the subword decreases, suggesting that frequent subwords require much less computation to encode. On further analysis we find that these subwords often correspond to function words. In contrast, we observe no such relationship between subword frequency and computation usage in the decoder. There are two exceptions to these phenomenon: (i) the special start-of-sentence token used in the encoder utilizes most of the available encoder compute, possibly playing a special role in encoding a sentence due to its presence in every training example, and (ii) the end-of-sentence token in the decoder needs very little computation to predict. The end-of-sentence token might be easier to predict due to the presence of punctuations preceding this token in most target sentences.

\paragraph{Layers are used adaptively when compute budget is limited:} We next attempt to understand how computation is allocated amongst different layers in the model when restricted to a limited budget. To understand the distribution of computation across layers at different budgets, we first analyze the fraction of inputs (subwords) for which each layer is active under different budget constraints over the entire validation set. Figure \ref{fig:capalloc} depicts the fraction of times self-attention, cross-attention and feed-forward layers are active in different encoder/decoder layers. We find that:

(i) Both encoders and decoders tend to utilize more feed-forward capacity higher up in the stack, i.e. feed-forward layers in higher encoder or decoder layers are active for a larger fraction of tokens. We found this to be true across multiple training runs. This also corroborates findings in previous work exploring where high capacity layers should be used in Transformer-based language models \citep{lample2019large}.
    
(ii) At very low budgets ($p=0.2$), 3 encoder self-attention layers, 2 decoder self-attention layers and 1 decoder cross-attention layer get completely pruned out. The position of these layers was different for different runs, for eg. while the 1st, 4th and 5th encoder self-attention layers got pruned out in the first run, the 1st, 3rd and 6th layer got pruned in the second. However, even at $p=0.2$, most layers are utilized adaptively depending on the input. At higher budgets ($p \in \{0.33, 0.5\}$) almost no layers are completely pruned.

\begin{figure*}[t]
  \centering
    \begin{subfigure}{0.66\columnwidth}
    \includegraphics[width=0.95\columnwidth]{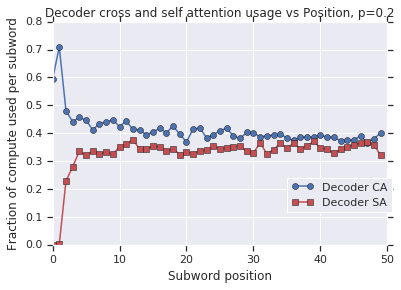}
    \end{subfigure}
    \begin{subfigure}{0.66\columnwidth}
    \includegraphics[width=0.95\columnwidth]{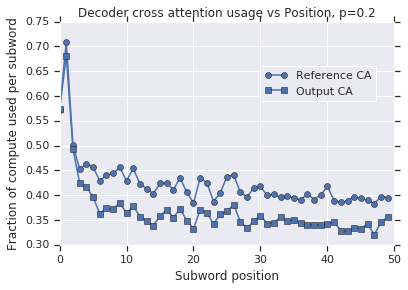}
    \end{subfigure}
    \begin{subfigure}{0.66\columnwidth}
    \includegraphics[width=0.95\columnwidth]{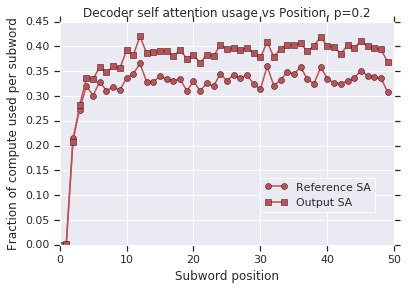}
    \end{subfigure}
  \caption{In these figures we compare the cross-attention and self-attention sub-network usage at different time-steps during decoding. The left plot depicts this comparison on the reference translations. The middle plot compares the cross-attention usage on reference translations against usage on  the model's decoder outputs, while the right plot compares self-attention usage on reference translations and the model's decoded outputs. The usage values are averaged over all the decoder layers.}
  \label{fig:savsca}
\end{figure*}

\begin{figure*}
  \centering
    \begin{subfigure}{0.9\columnwidth}
    \includegraphics[width=0.9\columnwidth]{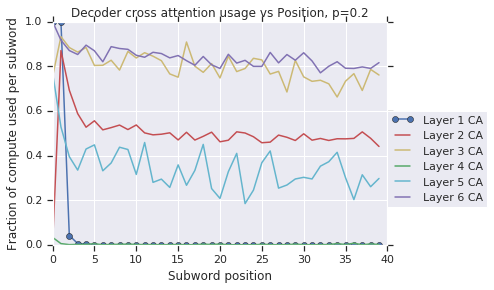}
    \end{subfigure}
    \begin{subfigure}{0.9\columnwidth}
    \includegraphics[width=0.9\columnwidth]{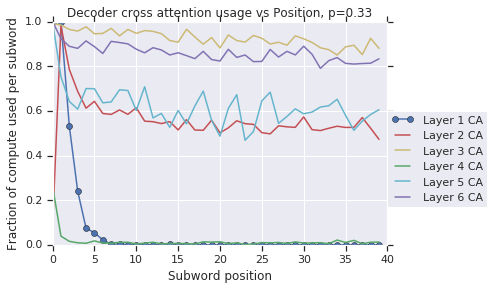}
    \end{subfigure}
  \caption{These figures depict the cross attention sub-network usage at different time-steps during auto-regressive decoding, for the cross-attention sub-network in all decoder layers, at $p=0.2$ (left plot) and $p=0.33$ (right plot).}
  \label{fig:caspec}
\end{figure*}
\begin{figure*}
  \centering
    \begin{subfigure}[b]{2.0\textwidth}
    \includegraphics[width=0.5\linewidth]{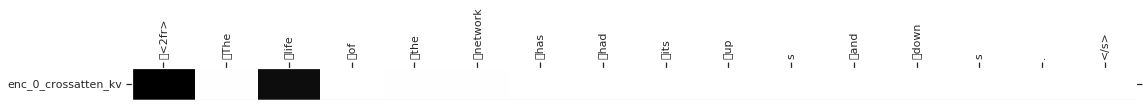}
    \end{subfigure}
    \begin{subfigure}[b]{2.0\textwidth}
    \includegraphics[width=0.5\linewidth]{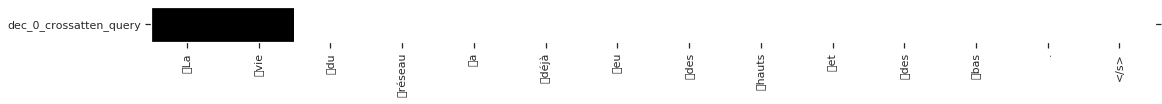}
    \end{subfigure}
  \caption{Image depicting the first decoder layer cross attention activations for a validation example. The top line depicts the sub-word fragmentation of the source sentence, followed by the cross-attention key-value subnetwork activation in the encoder. The bottom line depicts the target sub-word outputs and the cross-attention query subnetwork activation in the decoder.} 
  \label{fig:caspecex}
\end{figure*}

\paragraph{Decoder uses more cross-attention at early time-steps of decoding:} Given the nature of auto-regressive decoding, we would expect cross-attention and self-attention to be utilized differently as more target context becomes available. To understand the variation in self- and cross-attention usage over the decoding process, we evaluate the fraction of tokens each self- and cross-attention layer is active for, at different time-steps of decoding, averaged over all layers in the decoder. Figure~\ref{fig:savsca} depicts the results of this analysis. We notice that, at the beginning of decoding, the model learns to rely solely on information from the cross-attention sub-networks, which is understandable given that no self-attention context is available yet. As decoding progresses, the model allocates capacity almost equally between self- and cross-attention. This finding is consistent across multiple runs, and when the analysis is run on the model's decoded outputs instead of the reference translations. When comparing the fraction of self- and cross-attention usage on reference translations and model outputs, we notice a trend suggesting that the model might rely more on the self-attention sub-network when being used for decoding (with a corresponding reduction in cross-attention usage). This could be a result of exposure bias~\citep{bengio2015scheduled} forcing the model to rely more on self-attention since the distribution of the model outputs is different from the one it was trained on. This might also be explained by the model's outputs being samples of a smooth marginal distribution that is easier to learn (with much fewer modes), as compared against natural target text \citep{freitag2019text}.

\paragraph{First cross-attention layer is only used to predict the initial phrase in translation:} While analyzing control network outputs on different inputs, we noticed a very intriguing pattern of cross-attention usage in the first layer of the decoder. At low budgets ($p \in \{0.2, 0.33\}$), the 1st decoder cross-attention layer is only used when predicting the first few tokens. This phenomenon is illustrated in Figure~\ref{fig:caspec}, where we plot the usage of different cross-attention layers across decoder time-steps. At $p=0.2$, the 1st cross-attention layer is only used to predict the first 2-3 tokens of the target translation, while at $p=0.33$ this layer is active for the first 5-6 tokens. This pattern was consistent across multiple runs, and when evaluated on the model's decoded outputs. To further investigate this phenomenon we plot the cross-attention activity for both, the key-value sub-network acting on the encoder outputs, and the query sub-network acting on the decoder inputs on some samples from the validation set. One such example is depicted in Figure~\ref{fig:caspecex}, demonstrating that the key-value sub-network is active for the same token that translates into the first phrase of the target translation. This suggests that target word-order information is at-least partially determined in the encoder itself.

\section{Related Work}

Activating a sub-network depending on the particular input example has been the focus of conditional computation approaches \cite{Bengio2013EstimatingOP,davisA13}. Following this line of research, \citet{cho2014exponentially} studied increasing the capacity of neural networks without increasing required computation by exploiting the bit patterns associated with hidden units. As a majority of conditional computation approaches make use of stochastic binary units that pose trainability challenges, \citet{bengio2015conditional} cast the problem as a reinforcement learning problem and proposed a policy that maps the activations of layers to Bernoulli masks. \citet{graves2016adaptive} proposed the first application of conditional computation to neural sequence models, called Adaptive Computation Time (ACT), where a recurrent neural network is trained to learn the lag between reading an input and generating the output.

The recently introduced Transformer architecture \citep{vaswani2017attention} has allowed researchers to train neural networks with billions of parameters, reaffirming the need for more efficient and adaptive models. Universal transformer \citep{dehghani2018universal} addressed the parameter inefficiency problem of Transformers by tying the weights of consecutive layers and utilizing ACT to decide the halting of such recurrence. The recently proposed depth-adaptive Transformer (DAT) \cite{elbayad2019depthadaptive} is perhaps the most similar to our approach. In DAT, decoder layers are equipped with halting classifiers that decide to exit and predict the output or continue processing, extending the ACT framework. DAT requires explicit supervision from oracles (or implicit supervision from multiple softmax computations) to train halting classifiers, restricting their approach to specific applications (like decoders in sequence to sequence models). Our approach trains control networks end-to-end with the rest of the model, allowing us to extend it to a wider range of sub-networks not directly connected with the final classifier (for example, key-value projections in self-attention layers or encoder layers in seq2seq models).

\citet{fan2019reducing} propose another approach to control the inference time computation budget. Their method applies structured pruning (in the form of layer dropout), which allows selectively applying certain layers of a single network to control inference-time computation usage. Our approach however, results in a model that can simultaneously adapt to the difficulty of the input example and the computation budget that is available at hand during inference time. 

In addition, parallels can be made with approaches utilizing mixture-of-experts (MoE) \cite{10.1007/s10462-012-9338-y}, where different examples are routed to different experts in order to maximize the output diversity \cite{shen2019mixture} or device utilization \cite{shazeer2017outrageously}. 


\section{Conclusion}
\label{author info}

In this work we present a general framework to adapt neural sequence models (Transformer) for conditional computation and control the amount of computation used at inference. Our proposed approach injects simple control networks into the core computation graph, in order to modulate the information flow through the network. The incorporated control networks are trained end-to-end simultaneously with the model, simulating the binary decisions to be made at inference time. We also introduce a novel multi-task objective that allows the network to operate at multiple computation budgets at inference time efficiently, addressing the need for on-demand computation requirements of large networks. Experiments on large scale machine translation (WMT'14 English-French) and unsupervised representation learning (BERT) demonstrate that our proposed approach is competitive with baseline Transformer models at the same computation budget, and significantly better at smaller computational budgets compared to computationally equivalent baselines. 

Our analysis of the control network outputs reveals that the model learns to efficiently allocate capacity across inputs of different complexities, allowing it to function at reduced budgets without a significant drop in quality. 

\section*{Acknowledgements}
We would like to thank the Google Translate and Tensorflow Lingvo \citep{shen2019lingvo} teams for foundational contributions to the project.


\bibliography{cct}
\bibliographystyle{icml2020}



\end{document}